\documentclass[runningheads,a4paper]{llncs}

\usepackage{amsmath,amsfonts,amssymb}
\usepackage{graphicx}
\usepackage{algorithm}
\usepackage{algorithmicx}
\usepackage{algpseudocode}
\usepackage{multirow}
\usepackage{mathtools}
\usepackage{chngpage}
\usepackage{breqn}
\usepackage{xcolor}
\usepackage[normalem]{ulem}

\usepackage{hyperref}

\definecolor{dark-red}{rgb}{0.4,0.15,0.15}
\definecolor{dark-blue}{rgb}{0.15,0.15,0.4}
\definecolor{medium-blue}{rgb}{0.0,0.0,0.4}
\hypersetup{
    colorlinks, linkcolor={black},
    citecolor={dark-blue}, urlcolor={medium-blue}
}

\newcommand{\comment}[1]{}
\newcommand{\argmin}{\operatornamewithlimits{argmin}}
\newcommand{\argmax}{\operatornamewithlimits{argmax}}

\title{Discriminative Parameter Estimation\\
 for Random Walks Segmentation}

\author{
    Pierre-Yves~Baudin\inst{1-6} \and 
    Danny~Goodman\inst{1-3} \and 
    Puneet~Kumar\inst{1-3} \and 
    Noura~Azzabou\inst{4-6} \and
    Pierre~G.~Carlier\inst{4-6} \and 
    Nikos~Paragios\inst{1-3} \and
    M.~Pawan~Kumar\inst{1-3}
    }

\institute{
    Center for Visual Computing, \'{E}cole Centrale Paris, FR \and 
    Universit\'{e} Paris-Est, LIGM (UMR CNRS), \'{E}cole des Ponts ParisTech, FR \and 
    \'{E}quipe Galen, INRIA Saclay, FR \and
    Institute of Myology, Paris, FR \and 
    CEA, I$^{2}$ BM, MIRCen, IdM NMR Laboratory, Paris, FR \and 
    UPMC University Paris 06, Paris, FR
    }

\authorrunning{Pierre-Yves~Baudin et al.}

\begin{document}

\maketitle

\begin{abstract}
The Random Walks (RW) algorithm is one of the most efficient and easy-to-use
probabilistic segmentation methods. By combining contrast terms with prior
terms, it provides accurate segmentations of medical images in a fully automated manner.
However, one of the main drawbacks of using the RW algorithm is that its parameters have to be hand-tuned. \comment{In order to avoid this onerous task,} we propose a novel discriminative learning framework that estimates the parameters using a training dataset. The main challenge we face is that the training samples are not fully supervised. Specifically, they provide a hard segmentation of the \comment{medical} images, instead of a probabilistic segmentation. We overcome this challenge by treating the optimal probabilistic segmentation that is compatible with the given hard segmentation as a latent variable. This allows us to employ the latent support vector machine formulation for parameter estimation. We show that our approach significantly outperforms the baseline methods on a challenging dataset consisting of real clinical 3D MRI volumes of skeletal muscles.
\end{abstract}

\section{Introduction}
\label{sec:introduction}

The Random Walks (RW) algorithm is one of the most popular techniques for segmentation
in medical imaging~\cite{Grady2006}. Although it was initially proposed
for interactive settings,  
recent years have witnessed the development of fully automated extensions. In addition to the contrast information employed in the original formulation~\cite{Grady2006}, the automated extensions incorporate prior information based on appearance~\cite{Grady2005}
and shape~\cite{baudinMICCAI12}. 
% The state of the art RW algorithm has several benefits over its competitors: 
% (i) unlike surface-based methods like Active Contours, 
% it has the ability to handle intersection issues in multi-label cases; 
% (ii) it is robust to weak boundaries; and 
% (iii) it is computationally efficient due to the specific structure of the associated minimization problem. 

It has been empirically observed that the accuracy of the RW algorithm relies heavily on the relative weighting between the various contrast and prior terms. Henceforth, we refer to the relative weights of the various terms in the RW objective function as parameters. At present, researchers either rely on a user to hand-tune the parameters or on exhaustive cross-validation~\cite{baudinMICCAI12,Grady2005}. However, both these approaches quickly become infeasible as the number of terms in the RW objective function increase.

In contrast to the RW literature, the problem of parameter estimation has received considerable attention in the case of discrete models such as CRFs~\cite{Szummer2008}. Recent years have witnessed the emergence of structured-output support vector machine (Structured SVM) as one of the most effective discriminative frameworks for supervised parameter estimation~\cite{taskar2003maxmargin,tsochantaridis2004support}. Given a training dataset that consists of pairs of input and their ground-truth output, structured SVM minimizes the empirical risk of the inferred output with respect to the ground-truth output. The risk
is defined by a user-specified loss function that measures the difference in quality between two given outputs. 
% The structured SVM formulation has been successfully employed for many applications 
% including graph-cuts based segmentation~\cite{Szummer2008}.

% Inspired by the efficacy of structured SVM in discrete models, 
We would like to discriminatively learn the parameters of the RW formulation. To this end, a straightforward application
of structured SVM would require a training dataset that consists of pairs of inputs
% ---in our case, medical acquisitions such as MRI scans---
as well as their ground-truth outputs---in our case, the {\em optimal} probabilistic segmentation. In other words, we require a human to provide us with the output of the RW algorithm for the best set
of parameters. This is an unreasonable demand since the knowledge of the optimal probabilistic segmentation is as difficult to acquire as it is to hand-tune the parameters itself. Thus we cannot directly use structured SVM to estimate the desired parameters.

In order to handle the above difficulty, we propose a novel formulation for discriminative parameter estimation in the RW framework. Specifically, we learn the parameters using a weakly supervised dataset that consists of pairs of medical acquisitions and their hard segmentations. Unlike probabilistic segmentations, hard segmentations can be obtained easily from human annotators. We treat the optimal probabilistic segmentation that is {\em compatible} with the hard segmentation as a latent variable. Here, compatibility refers to the fact that the probability of the ground-truth label (as specified by the hard segmentation) should be greater than the probability of all other labels for each pixel/voxel. The resulting representation allows us to learn the parameters using the latent SVM formulation~\cite{felzenszwalb2008discriminatively,smola2005kernel,Yu2009}.
% Note that, while several other semi-supervised learning models 
% have been proposed in the literature~\cite{zhu2003semi,zhu2005semi}, 
% we choose to build on latent SVM as it allows us to directly minimize an
% appropriate regularized risk function over the training dataset.

While latent SVM does not result in a convex optimization problem, its local optimum solution can be obtained using the iterative concave-convex procedure (CCCP)~\cite{yuille2003concave}. 
% At each iteration, CCCP performs two steps: 
% (i) estimating a compatible probabilistic segmentation for each training sample using the current
% set of parameters---commonly referred to as annotation consistent inference (ACI); 
% and (ii) updating the parameters by fixing the compatible probabilistic 
% segmentations to those obtained during ACI. 
% The second step of CCCP involves solving a structured SVM
The CCCP involves solving a structured SVM
 problem, which lends itself to efficient optimization. In order to make the overall algorithm computationally feasible, we propose a novel efficient approach for ACI based on dual decomposition~\cite{bertsekas99,Komodakis2007}. We demonstrate the benefit of our learning framework over a baseline structured SVM using a challenging dataset of real 3D MRI volumes.

\section{Preliminaries}
\label{sec:preliminaries}
We will assume that the input ${\bf x}$ is a 3D volume. We denote the $i$-th voxel of ${\bf x}$ as ${\bf x}(i)$, and the set of all voxels as ${\cal V}$. In a hard segmentation, each voxel is assigned a label $s \in {\cal S}$ (for example, the index of
a muscle). We will use ${\bf z}$ to represent the human annotation (that is, the class labels of the voxels in ${\bf x}$) in binary form: 
\begin{equation}
\small
{\bf z}\left(i,s\right)=\begin{cases}
1 & \mbox{ if voxel $i \in {\cal V}$ is of class $s \in {\cal S}$},\\
0 & \mbox{ otherwise}.
\end{cases}
\end{equation}
In other words, the binary form ${\bf z}$ of the annotation specifies delta distribution over the putative labels for
each voxel. Our training dataset is a collection of training images
${\bf x}$ and hard segmentations ${\bf z}$: $\mathcal{D}=\{\left({\bf x}_{k},{\bf z}_{k}\right)\}_{k}$.
Note that we use subscript $k$ to denote the input index within a
dataset, and parenthetical $i$ to denote a voxel within a particular
input.

\subsection{Random Walks Segmentation}
The RW algorithm provides a probabilistic---or soft---segmentation of an input ${\bf x}$, which we denote by ${\bf y}$, that is,
\begin{equation}
\small
{\bf y}(i,s)=\mathrm{Pr}\left[\mbox{voxel \ensuremath{i} is of class \ensuremath{s}}\right], \forall i \in {\cal V}, s \in {\cal S}\,.
\end{equation}
When using one contrast term and one prior model, the RW algorithm
amounts to minimizing the following convex quadratic objective functional:
{\small
\begin{align}
E\left({\bf y},{\bf x}\right) & = 
   {\bf y}^{\top}L\left({\bf x}\right){\bf y} + 
       w^{\mathrm{prior}}\left\Vert {\bf y}-{\bf y}_{0}\right\Vert_{\Omega_0({\bf x})}^2  \,,\\
 & = {\bf y}^{\top}L\left({\bf x}\right){\bf y}+E^{\mathrm{prior}}({\bf y},{\bf x})  \,.
\label{eq:RW+prior}
\end{align}}%
Here, ${\bf y}_{0}$ is a reference prior probabilistic segmentation dependent on appearance~\cite{Grady2005} or
shape~\cite{baudinMICCAI12}, and $\Omega_0({\bf x})$ is a diagonal matrix that specifies a
voxel-wise weighting scheme for ${\bf x}$. The term $L({\bf x})$ refers to a combinatorial Laplacian matrix defined on a neighborhood system ${\cal N}$ based on the adjacency of the voxels. It is a block diagonal matrix---one block per label---with all identical blocks, where the entries of the block $L^b({\bf x}) $ use the typical Gaussian kernel formulation (see ~\cite{Grady2006}). The relative weight $w^{\mathrm{prior}}$ is the parameter
for the above RW framework. The above problem is convex, and can be optimized efficiently by solving a sparse linear system of equations. We refer the reader
to~\cite{baudinMICCAI12,Grady2006} for further details.

\subsection{Parameters and Feature Vectors}
In the above description of the RW algorithm, we restricted ourselves to a single Laplacian and a single prior.
However, our goal is to enable the use of numerous Laplacians and priors. To this end, let
$\{L_{\alpha}\}_{\alpha}$ denote a known family of Laplacian matrices and
$\left\{ E_{\beta}\left(\cdot\right)\right\} _{\beta}$ denote a known family of
prior energy functionals. In section~\ref{sec:experiments}, we will specify the
family of Laplacians and priors used in our experiments. We denote the general form
of a linear combination of Laplacians and prior terms as:
{\small
\begin{align}
L\left({\bf x};{\bf w}\right)  =  \sum_{\alpha}w_{\alpha}L_{\alpha}\left({\bf x}\right),
E^{\mathrm{prior}}\left(\cdot,{\bf x};{\bf w}\right)  =  \sum_{\beta}w_{\beta}E_{\beta}\left(\cdot,{\bf x}\right), {\bf w} \geq 0  \,.
\end{align}
}%
Each term $E_{\beta}\left(\cdot,{\bf x}\right)$ is of the form:
\begin{equation}
\small
E_{\beta}\left({\bf y},{\bf x}\right)=\left\Vert {\bf y}-{\bf y}_{\beta}\right\Vert _{\Omega_{\beta}\left({\bf x}\right)}^{2}  \,,
\end{equation}
where ${\bf y}_{\beta}$ is the $\beta$-th reference segmentation and $\Omega_{\beta}({\bf x})$ is the corresponding
voxel-wise weighting matrix (which are both known). We denote the set of all parameters as
${\bf w}=\left\{ w_{\alpha},w_{\beta}\right\} _{\alpha,\beta}$. Clearly,
the RW energy \eqref{eq:RW+prior} is linear in ${\bf w}$, and can therefore
be formulated as:
{\small
\begin{align}
E\left({\bf y},{\bf x};{\bf w}\right) & =
    {\bf y}^{T}L\left({\bf x};{\bf w}\right){\bf y}
        + E_{\mathrm{prior}}\left({\bf y},{\bf x};{\bf w}\right)  \,,\\
    & = {\bf w}^{T}\psi\left({\bf x},{\bf y}\right)   \,,    \label{eq:RW_psi}
\end{align}
}%
where $\psi\left({\bf x},{\bf y}\right)$ is known as the joint feature
vector of ${\bf x}$ and ${\bf y}$. Note that by restricting the parameters to be non-negative (that is, ${\bf w} \geq {\bf 0}$),
we ensure that the
energy functional $E(\cdot,{\bf x};{\bf w})$ remains convex.

\subsection{Loss Function}
As mentioned earlier, we would like to estimate the parameters ${\bf w}$
by minimizing the empirical risk over the training samples. The risk
is specified using a loss function that measures the difference
between two segmentations. In this work, we define the loss function as the number of
incorrectly labeled voxels. Formally, let $\hat{{\bf y}}$ denote the underlying hard segmentation of the
soft segmentation ${\bf y}$, that is,
$\hat{{\bf y}}\left(i,s\right)=\delta\left(s=\argmax_{s \in {\cal S}}{\bf y}\left(i,s\right)\right)$,
where $\delta$ is the Kronecker function. The loss function is defined as
{\small
\begin{align}
\Delta\left({\bf z},{\bf y}\right) & = 1-\frac{1}{|{\cal V}|}\,\hat{{\bf y}}^{T}{\bf z}   \,,
\label{eq:loss}
\end{align}
}%
where ${\cal V}$ is the set of all voxels, and $|\cdot|$ denotes the cardinality of a set.

\section{Parameter Estimation Using Latent SVM}
\label{sec:parametere stimation using LSVM}
Given a dataset ${\cal D} = \{({\bf x}_k,{\bf z}_k),k=1,\cdots,N\}$, which consists of inputs ${\bf x}_k$ and their
hard segmentation ${\bf z}_k$, we would like to estimate parameters ${\bf w}$ such that the resulting inferred
segmentations are accurate.
Here, the accuracy is measured using the loss function $\Delta(\cdot,\cdot)$.
Formally, let ${\bf y}_k({\bf w})$ denote the soft segmentation obtained by minimizing the energy functional
$E(\cdot,{\bf x}_k;{\bf w})$ for the $k$-th training sample, that is,
\begin{equation}
\small
{\bf y}_k({\bf w}) = \argmin_{{\bf y}} {\bf w}^\top\psi({\bf x}_k,{\bf y})   \,.
\label{eq:prediction}
\end{equation}
We would like to learn the parameters ${\bf w}$ such that the empirical
risk is minimized. In other words, we would like to estimate the parameters ${\bf w}^*$ such that
\begin{equation}
\small
{\bf w}^* = \argmin_{{\bf w}} {\frac{1}{N} \sum_k \Delta({\bf z}_k,{\bf y}_k({\bf w}))}   \,.
\end{equation}
The above objective function is highly non-convex in ${\bf w}$, which makes it prone to bad local minimum solutions.
To alleviate this deficiency, it can be shown that the following latent SVM formulation minimizes a regularized upper bound on the risk for a 
set of samples $\{({\bf x}_k,{\bf z}_k),k=1,\cdots,N\}$:
{\small
\begin{align}
\label{eq:latentSVM}
\min_{{\bf w} \geq {\bf 0}} &
     \lambda ||{\bf w}||^2 + \lambda' ||{\bf w}-{\bf w}_0||^2 + \frac{1}{N} \sum_k \xi_k  \,, \\
\mbox{s.t.} &
     \min_{{\bf y}_k,\Delta({\bf z}_k,{\bf y}_k) = 0} {\bf w}^\top\psi({\bf x}_k,{\bf y}_k)
\leq {\bf w}^\top\psi({\bf x}_k,\overline{{\bf y}}_k) - \Delta({\bf z}_k,\overline{{\bf y}}_k) + \xi_k  \,,
    \forall \overline{{\bf y}}_k, \forall k  \,, \nonumber
\end{align}
}%
where the slack variable $\xi_k$ represents the upper bound of the risk for the $k$-th training sample.
Note that we have added two regularization terms for the parameters ${\bf w}$. The first term $||{\bf w}||^2$, weighted by hyperparameter
$\lambda$, ensures that we do not overfit to the training samples. The second term $||{\bf w}-{\bf w}_0||^2$, weighted by hyperparameter
$\lambda'$, ensures that we do not obtain a solution that is very far away from our initial estimate ${\bf w}_0$. The reason for including
this term is that our upper bound to the empirical risk may not be sufficiently tight. Thus, if we do not
encourage our solution to lie close to the initial estimate, it may drift towards an inaccurate set of parameters.
In section~\ref{sec:experiments}, we show the empirical
effect of the hyperparameters $\lambda$ and $\lambda'$ on the accuracy of the parameters.

While the upper bound of the empirical risk derived above is not convex, it was shown to be a difference of two convex
functions in~\cite{Yu2009}. This observation allows us to obtain a local minimum or saddle point solution using the
CCCP algorithm~\cite{Yu2009,yuille2003concave}, outlined in Algorithm~\ref{algo:CCCP}, which iteratively improves the parameters starting with an initial estimate 
${\bf w}_0$. It consists of two main steps at each iteration: (i) step 3, which involves estimating a compatible
soft segmentation for each training sample---known as annotation consistent inference (ACI); and (ii) step 4, which involves updating
the parameters by solving problem~(\ref{eq:structSVM}). In the following subsections, we provide efficient algorithms for both the
steps.

\begin{algorithm}
\small
\caption{\label{algo:CCCP} The CCCP method for parameter estimation using latent SVM.}
\textbf{Input:} Dataset $\mathcal{D}$, $\lambda$, $\lambda'$, ${\bf w}_0$, $\varepsilon$
\begin{algorithmic}[1]
\State Set $t=0$. Initialize ${\bf w}_t = {\bf w}_0$.
\Repeat 
\State Compute ${\bf y}_k^* = \argmin_{{\bf y}_k,\Delta({\bf z}_k,{\bf y}_k) = 0} {\bf w}_t^\top\psi({\bf x}_k,{\bf y}_k), \forall k$.
\State Update the parameters by solving the following problem
{\small
\begin{align}
\label{eq:structSVM}
{\bf w}_{t+1} = \argmin_{{\bf w} \geq {\bf 0}} & \lambda ||{\bf w}||^2 + \lambda' ||{\bf w}-{\bf w}_0||^2 + \frac{1}{N} \sum_k \xi_k   \,, \\
\mbox{s.t.} & {\bf w}^\top\psi({\bf x}_k,{\bf y}_k^*)
\leq {\bf w}^\top\psi({\bf x}_k,\overline{{\bf y}}_k) - \Delta({\bf z}_k,\overline{{\bf y}}_k) + \xi_k,
    \forall \overline{{\bf y}}_k, \forall k    \,, \nonumber
\end{align}
}%
\State $t=t+1$
\Until{The objective function of problem~(\ref{eq:latentSVM}) does not decrease below tolerance $\varepsilon$.}
\end{algorithmic}
\end{algorithm}

\subsection{Annotation Consistent Inference}
Given an input ${\bf x}$ and its hard segmentation ${\bf z}$, ACI requires us to find the soft segmentation ${\bf y}$ with the minimum energy, under the
constraint that it should be compatible with ${\bf z}$ (see step 3 of Algorithm~\ref{algo:CCCP}). We denote the ground truth label of a voxel $i$ by
$s_i$, that is, $s_i = \argmax_s {\bf z}(i,s)$, and the set of all voxels by ${\cal V}$. Using our notation, ACI can be formally specified as
\begin{equation}
\small
\min_{{\bf y} \in {\cal C}({\cal V})} {\bf y}^\top L({\bf x};{\bf w}) {\bf y} + E^{\mathrm{prior}}({\bf y},{\bf x};{\bf w})   \,.
\label{eq:ACI}
\end{equation}
Here, ${\cal C}({\cal V})$ is the set of all compatible probabilistic segmentations, that is,
{\small
\begin{align}
\label{eq:nonNegative}
& {\bf y}(i,s) \geq 0, \forall i \in {\cal V}, \forall s \in {\cal S}   \,, \\
\label{eq:normalization}
& \sum_{s \in {\cal S}} {\bf y}(i,s) = 1, \forall i \in {\cal V}   \,, \\
\label{eq:compatibility}
& {\bf y}(i,s_i) \geq {\bf y}(i,s), \forall i \in {\cal V}, \forall s \in {\cal S}   \,.
\end{align}
}%
Constraints~(\ref{eq:nonNegative}) and~(\ref{eq:normalization}) ensure that ${\bf y}$ is a valid probabilistic segmentation. The last
set of constraints~(\ref{eq:compatibility}) ensure that ${\bf y}$ is compatible with ${\bf z}$. Note that in the absence of
constraints~(\ref{eq:compatibility}), the above problem can be solved efficiently using the RW algorithm. However, since the ACI problem
requires the additional set of compatibility constraints, we need to develop a novel efficient algorithm to solve the above convex
optimization problem. To this end, we exploit the powerful dual decomposition framework~\cite{bertsekas99,Komodakis2007}. 
Briefly, we divide the above problem into a set of smaller subproblems defined using overlapping subsets of variables. Each subproblem can
be solved efficiently using a standard convex optimization package. In order to obtain the globally optimal solution of the original subproblem, we
pass {\em messages} between subproblems until they agree on the value of all the shared variables.
For details on the ACI algorithm, please refer to the supplementary materials of this paper.

\subsection{Parameter Update}
Having generated a compatible soft segmentation, the parameters can now be efficiently updated by solving problem~(\ref{eq:structSVM}) for a fixed set of
soft segmentations ${\bf y}_k^*$. 
This problem can be solved efficiently using the popular cutting plane method (for details on this algorithm, please refer to~\cite{Joachims2009}). Briefly, the method starts
by specifying no constraints for any of the training samples.
At each iteration, it finds the most violated constraint for each sample,
and updates the parameters until the increase in the objective function is less than a small epsilon.

In this work, due to the fact that our loss function is not concave, we approximate
the most violated constraint as the predicted segmentation, that is,
\begin{equation}
\small
\overline{{\bf y}} = \argmin_{{\bf y}} {\bf w}^\top\psi({\bf x},{\bf y})   \,.
\end{equation}
The above problem is solved efficiently using the RW algorithm.

\section{Experiments}
\label{sec:experiments}
\begin{figure}[t]%placement
    \centering
    \includegraphics[width=1\textwidth]{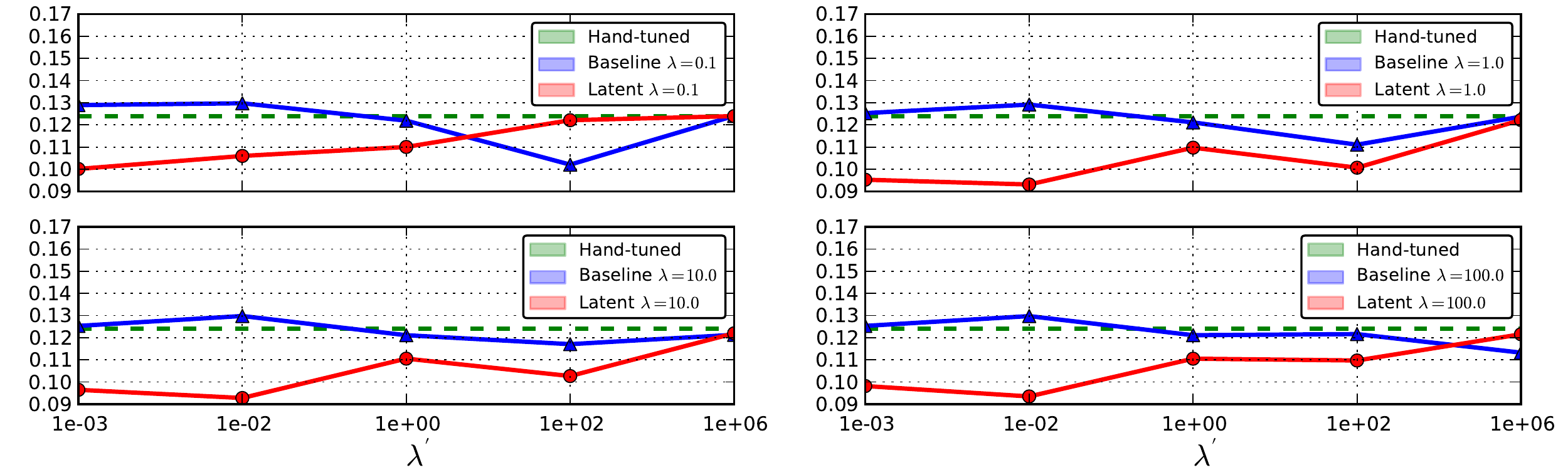}
    % \vspace{-1em}
    \caption{Estimated risk $\Delta(\mathbf{y}^{\star}_k, \mathbf{y}_k(\mathbf{w}))$ for three different methods}
    \label{fig:loss}
    % \vspace{-1em}
\end{figure}  
\emph{Dataset.} The dataset consists of 30 MRI
volumes of the thigh region of dimensions $224\times 224\times 100$.
The various segments correspond to 4 different muscle groups together with the background class. We randomly split the dataset into 80\% for
training and 20\% for testing. In order to reduce the training time for both our method and the baselines, we divide each volume
into $100/2$ volumes of dimension $224\times 224\times 2$. 

\emph{Laplacians and Prior Terms.} We use 4 different Laplacians (generated with different weitghing functions). Furthermore, we use
two shape priors based on~\cite{baudinMICCAI12} and one appearance prior based on~\cite{Grady2005}.  This results in a total of 7 parameters to be estimated. 

\emph{Methods.} The main hypothesis of our work is that it is important to represent the unknown optimal soft segmentation using latent variables. Thus we compare our method with a baseline structured SVM that replaces
the latent variables with the given hard segmentations. In other words, our baseline estimates the parameters by solving problem~(\ref{eq:structSVM}), where the imputed soft segmentations ${\bf y}^*_k$ are
replaced by the hard segmentations ${\bf z}_k$. During our experiments, we found that replacing the hard segmentation with a pseudo soft segmentation
based on the distance transform systematically decreased the loss of the output. Thus the method refered to as "Baseline" uses a structured SVM with distance-tranform "softened" segmentations. 

\begin{figure}
    \centering
    \includegraphics[trim=10em 30em 10em 13em, clip=true, width=0.24\textwidth]{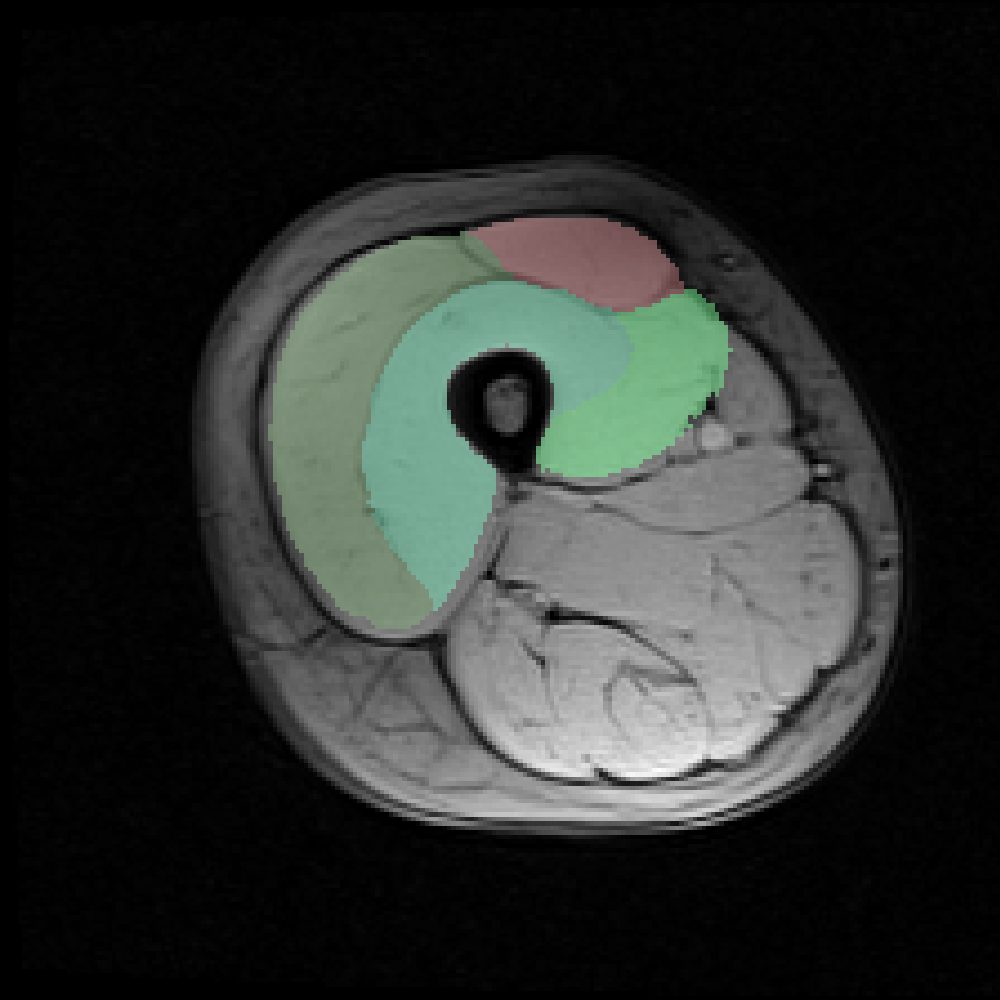}
    \includegraphics[trim=10em 30em 10em 13em, clip=true, width=0.24\textwidth]{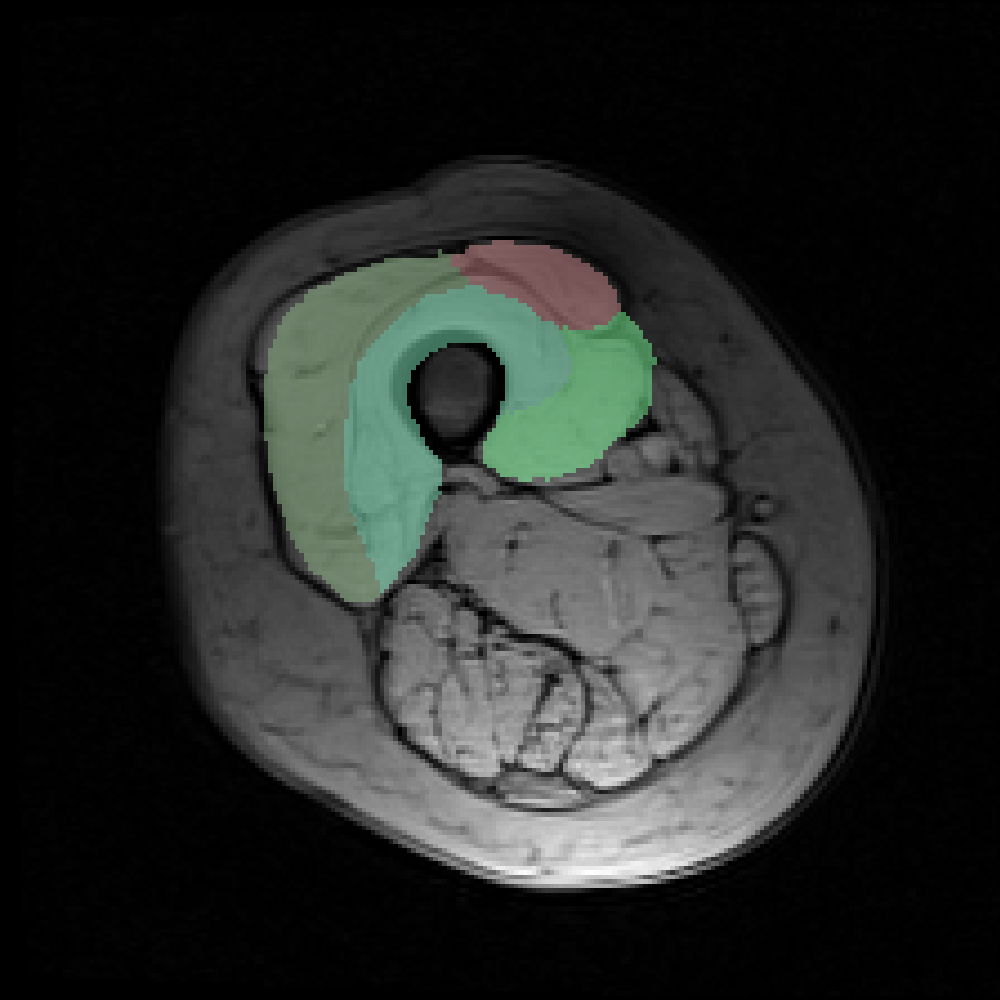}
    \includegraphics[trim=10em 30em 10em 13em, clip=true, width=0.24\textwidth]{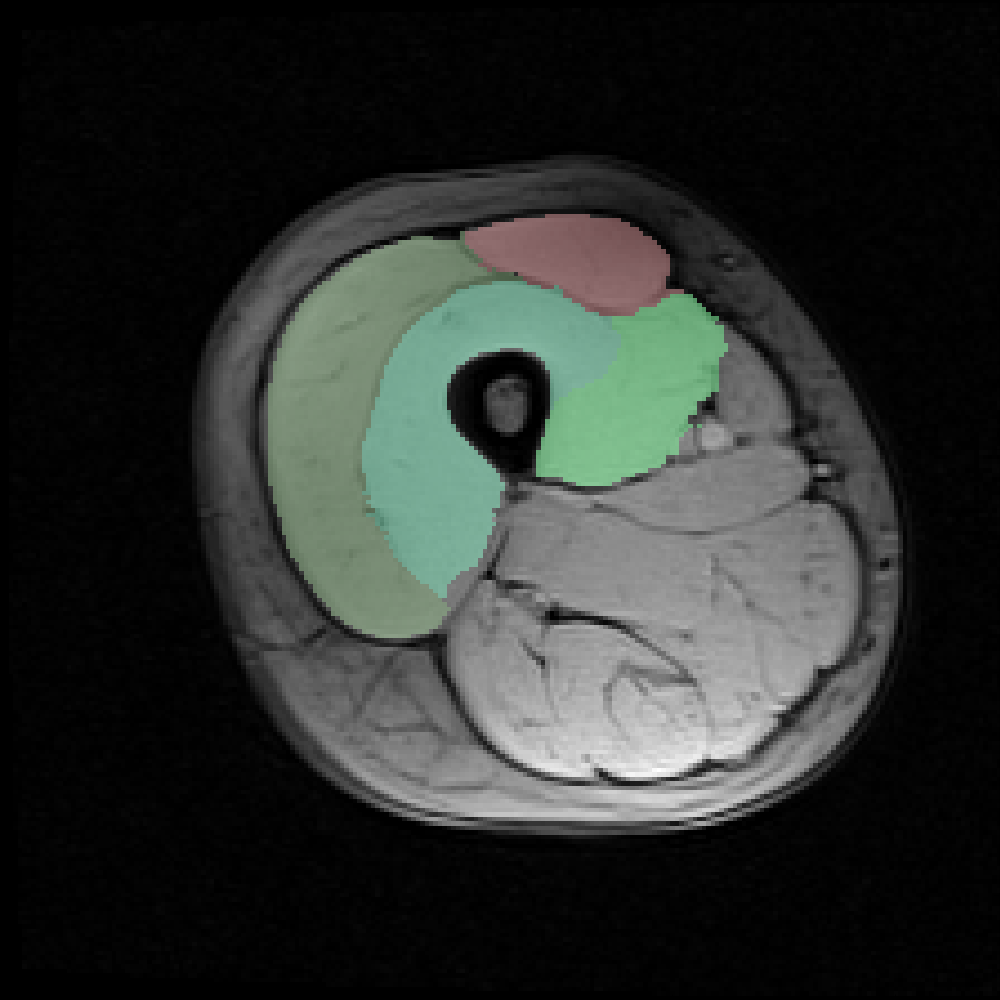}
    \includegraphics[trim=10em 30em 10em 13em, clip=true, width=0.24\textwidth]{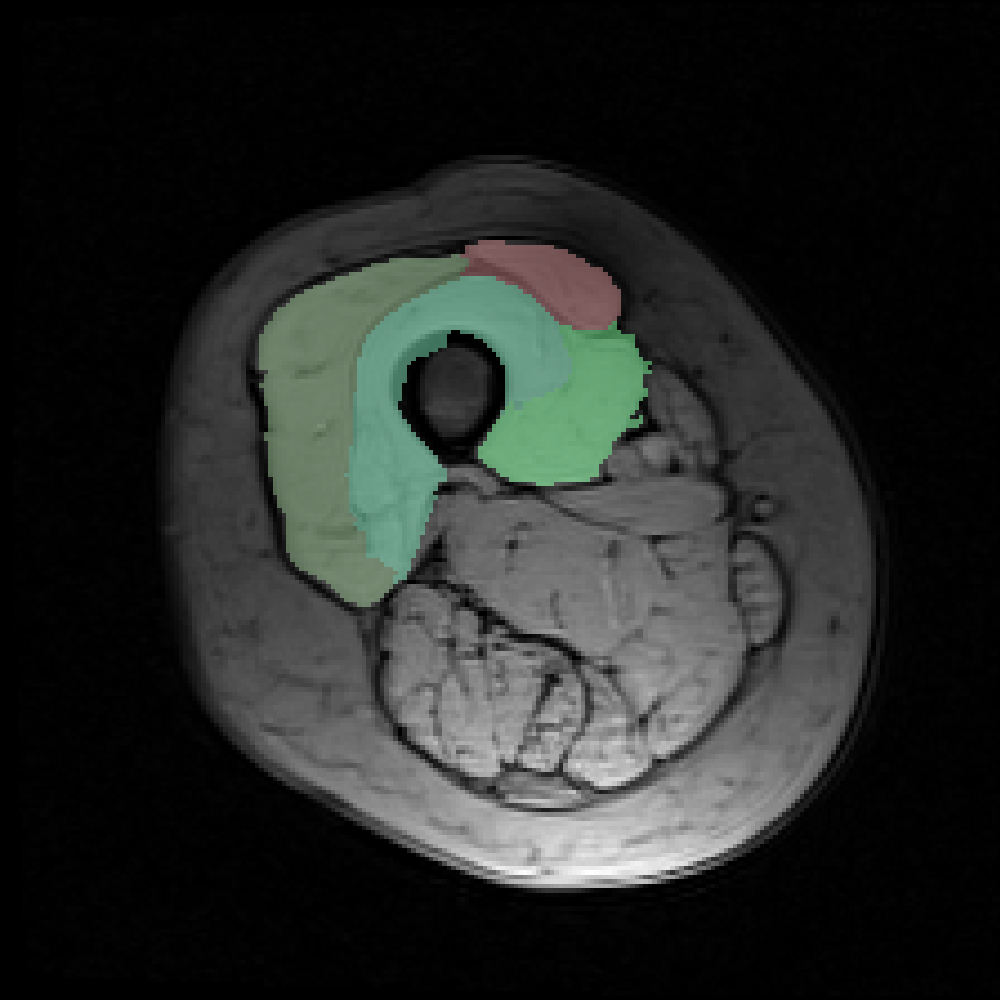}
    \caption{Method comparison: (columns 1 \& 2) segmentations using $\bf{w}_0$; (columns 3 \& 4) segmentations using learned $\bf{w}$ using latent structured SVM. The latter are closer to expert segmentation.}
    \label{fig:cross}
    % \vspace{-1.5em}
    \end{figure}

\emph{Results.} Fig.~\ref{fig:loss} shows the test loss for three different methods: (i) the initial hand-tuned parameters ${\bf w}_0$;
(ii) the baseline structured SVM with distance transforms; and (iii) our proposed approach using latent SVM. 
As can be seen from Fig.~\ref{fig:loss}, latent SVM provides significantly better results than the baselines---even when using the distance transform. For the 4 x 5 hyperparameter settings that we report (that is, four different values of $\lambda$ and 5 different values of $\lambda^\prime$), latent SVM is significantly better than SVM in 15 cases, and significantly worse in only 2 cases.
Note that latent SVM provides the best results for very small values of $\lambda^\prime$, which indicates that the upper bound on the empirical risk in tight. As expected, for sufficiently large values of $\lambda'$, all the methods provide similar results. For the best settings of the corresponding hyperparameters, the
percentage of incorrectly labeled voxels as follows: (i) for ${\bf w}_0$, $13.5\%$; (ii) for structured SVM, $10.0\%$; and
(iii) for latent SVM, $9.2\%$. Fig.~\ref{fig:cross} shows some example segmentations for the various methods.  

\section{Discussion}
We proposed a novel discriminative learning framework to estimate the parameters for the probabilistic RW segmentation algorithm.
We represented the optimal soft segmentation that is compatible with the hard
segmentation of each training sample as a latent variable. This allowed us to formulate the problem of parameter estimation using latent
SVM, which upper bounds the empirical risk of prediction with a difference of convex optimization program. Using a challenging clinical
dataset of MRI volumes, we demonstrated the efficacy of our approach over the baseline method that replaces the latent variables with
the given hard segmentations. 
The latent SVM framework can be used to estimate parameters with partial hard segmentations. Such an approach would allow us to
scale the size of the training dataset by orders of magnitude. 
% \vspace{-1em}
\bibliographystyle{splncs03}
\bibliography{learnRW}

\begin{thebibliography}{10}
\providecommand{\url}[1]{\texttt{#1}}
\providecommand{\urlprefix}{URL }

\bibitem{baudinMICCAI12}
Baudin, P.Y., Azzabou, N., Carlier, P., Paragios, N.: Prior knowledge, random
  walks and human skeletal muscle segmentation. In: MICCAI (2012)

\bibitem{bertsekas99}
Bertsekas, D.: Nonlinear Programming. Athena Scientific (1999)

\bibitem{felzenszwalb2008discriminatively}
Felzenszwalb, P., McAllester, D., Ramanan, D.: A discriminatively trained,
  multiscale, deformable part model. In: CVPR (2008)

\bibitem{Grady2005}
Grady, L.: Multilabel random walker image segmentation using prior models. In:
  CVPR (2005)

\bibitem{Grady2006}
Grady, L.: Random walks for image segmentation. PAMI  (2006)

\bibitem{Joachims2009}
Joachims, T., Finley, T., Yu, C.N.: Cutting-plane training of structural
  {SVMs}. Machine Learning  (2009)

\bibitem{Komodakis2007}
Komodakis, N., Paragios, N., Tziritas, G.: {MRF} optimization via dual
  decomposition: Message-passing revisited. In: ICCV (2007)

\bibitem{smola2005kernel}
Smola, A., Vishwanathan, S., Hoffman, T.: Kernel methods for missing variables.
  In: AISTATS (2005)

\bibitem{Szummer2008}
Szummer, M., Kohli, P., Hoiem, D.: Learning {CRFs} using graph cuts. In: ECCV
  (2008)

\bibitem{taskar2003maxmargin}
Taskar, B., Guestrin, C., Koller, D.: Max-margin {M}arkov networks. In: NIPS
  (2003)

\bibitem{tsochantaridis2004support}
Tsochantaridis, I., Hofmann, T., Joachims, T., Altun, Y.: Support vector
  machine learning for interdependent and structured output spaces. In: ICML
  (2004)

\bibitem{Yu2009}
Yu, C.N., Joachims, T.: Learning structural {SVMs} with latent variables. In:
  ICML (2009)

\bibitem{yuille2003concave}
Yuille, A., Rangarajan, A.: The concave-convex procedure ({CCCP}). Neural
  Computation  (2003)

\end{thebibliography}

\end{document}